# MEME-Fusion@CHiPSAL 2026: Multimodal Ablation Study of Hate Detection and Sentiment Analysis on Nepali Memes

**Samir Wagle**\*, **Reewaj Khanal**\*, **and Abiral Adhikari**\*
Department of Computer Science and Engineering, Kathmandu University, Nepal
{waglesameer5, emailreewaj, abiraladhikari1222}@gmail.com
\* These authors have contributed equally.

**Abstract**

Hate speech detection in Devanagari-scripted social media memes presents compounded challenges: multimodal content structure, script-specific linguistic complexity, and extreme data scarcity in low-resource settings. This paper presents our system for the CHiPSAL 2026 shared task, addressing both Subtask A (binary hate speech detection) and Subtask B (three-class sentiment classification: positive, neutral, negative). We propose a hybrid cross-modal attention fusion architecture that combines CLIP (ViT-B/32) for visual encoding with BGE-M3 for multilingual text representation, connected through 4-head self-attention and a learnable gating network that dynamically weights modality contributions on a per-sample basis. Systematic evaluation across eight model configurations demonstrates that explicit cross-modal reasoning achieves a 5.9% F1-macro improvement over text-only baselines on Subtask A, while uncovering two unexpected but critical findings: English-centric vision models exhibit near-random performance on Devanagari script, and standard ensemble methods catastrophically degrade under data scarcity (N ≈ 850 per fold) due to correlated overfitting. The code can be accessed at
https://github.com/Tri-Yantra-Technologies/MEME-Fusion/

## 1. Introduction

The proliferation of hate speech on social media platforms has become a critical concern in the moderation of online content, with particular severity in multilingual and multimodal contexts. While English text-based hate speech has received substantial research attention, non-English languages employing non-Latin scripts such as Devanagari remain significantly underserved. This gap is consequential: Hindi, Nepali, Marathi, and related Devanagari-scripted languages collectively serve over 600 million speakers who increasingly engage with social media through meme formats, a modality that fuses visual and textual signals into a semantically unified artifact. The CHiPSAL 2026 shared task Thapa et al. (2026), Sarveswaran et al. (2026) on Multimodal Hate and Sentiment Understanding in Low-Resource Text-Embedded Images directly addresses this challenge. Our system participates in two subtasks: Subtask A (binary hate speech detection: hate vs. non-hate) and Subtask B (three-class sentiment classification: positive, neutral, negative). Both subtasks operate on a curated corpus of 1,068 Devanagari meme instances, requiring joint reasoning over OCR-extracted text and meme image content. Three structural challenges characterize this problem. First, the dataset is severely limited in scale (N = 1,068 total), creating an acute sample-to-parameter scarcity that destabilizes high-dimensional fusion strategies. Second, class imbalance is pronounced in Subtask A, with hate instances constituting 67.4% of the corpus. Third, standard pretrained vision models such as CLIP were predominantly trained on English web data, fundamentally limiting their representational capacity for Devanagari script and culturally specific Nepali visual content. To address these challenges, we propose a hybrid cross-modal attention fusion architecture. At its core, 4-head self-attention computes explicit dependencies between visual embeddings from CLIP and textual embeddings from BGE-M3, followed by a learnable softmax-normalized gating network that assigns per-sample modality weights. This allows the model to privilege text when linguistic hate signals are explicit (e.g., direct slurs) and to upweight visual features when textual content is ambiguous. Critically, our ablation studies reveal that naive approaches: simple concatenation and traditional ensemble methods actively harm performance in this low-resource regime, a finding with direct practical consequences for NLP system design in data-scarce settings.

## 2. Related Work

The research trajectory for computational hate speech detection within Devanagari-scripted languages has been predominantly catalyzed by the Hate Speech and Offensive Content (HASOC) shared task series, which served to establish crucial benchmark datasets and standardized evaluation protocols, primarily for Hindi Mandl et al. (2019).

The methodological approaches ranging from classical feature engineering, such as Term Frequency-Inverse Document Frequency (TF-IDF) and character n-grams, to sophisticated transformer-based architectures, such as multilingual BERT variants are continuously evolving. Although these deep learning models consistently demonstrate superior performance over conventional baselines, these models are less effective when computational resources are limited. When fine tuning data is scarce, the model tend to provide English-centric biases that can be mitigated by more robust datasets. This evolving architectural landscape is further complemented by earlier deep learning efforts, notably the Bi-directional Long Short-Term Memory (Bi-LSTM) framework Adhikari et al. (2024), which successfully delineated between profanity and general offensiveness in Nepali and recorded a high-watermark accuracy of 87.8% on an analogously data-constrained corpus. More cutting-edge investigations are now focused on leveraging Large Language Model (LLM)-based paradigms, employing zero-shot and few-shot learning for South Asian content moderation Thapa et al. (2025a), as well as exploring retrieval-augmented distillation strategies to strategically transfer and exploit knowledge representations acquired from data-rich languages Thapa et al. (2025b).

The extension from text-only to multimodal hate detection reflects the growing predominance of meme-based communication on social media. Bhandari et al. (2023) introduced the CrisisHateMM dataset focusing on hate speech in text-embedded images from geopolitical conflict contexts, demonstrating that joint vision-language models outperform unimodal baselines by 8–12% F1. Thapa et al. (2025a) further explored prompt-based frameworks for code-mixed and low-resource meme understanding, showing that carefully engineered prompt templates improve model performance even under data scarcity. These works primarily evaluate fusion strategies on datasets with thousands of labeled instances; their applicability to extreme low-resource regimes (N < 1,100) remains an open question addressed by our study.

OpenAI's CLIP model Radford et al. (2021), pre-trained on 400 million image-text pairs from the English web, provides strong zero-shot generalization for Latin-script and Western visual content. On top of this multilingual vision-language pre-training(mCLIP, IndicCLIP) was carried out to prevent performance degradation on non-Latin scripts. However, Devanagari coverage in these models still remains limited. For text encoding, the BAAI General Embedding model BGE-M3 Xiao et al. (2023) offers strong multilingual dense retrieval performance with explicit support for Indic scripts, outperforming mBERT and XLM-RoBERTa on low-resource language retrieval benchmarks. The complementarity of CLIP's visual capabilities and BGE-M3's multilingual text understanding motivates our dual-encoder architecture.

## 3. Dataset, Tasks, and Preprocessing

### 3.1. Task Definition

The CHiPSAL 2026 shared task provides a corpus of Devanagari meme instances, each consisting of a JPEG image file containing visual content with embedded Devanagari text, and a text field containing OCR-extracted Devanagari/Nepali text. Our system addresses two subtasks:

- Subtask A – Binary Hate Speech Detection: Given a meme, classify it as Hate (1) or Non-Hate (0).

- Subtask B – Sentiment Classification: Given a meme, classify it as Positive (2), Neutral (1), or Negative (0).

### 3.2. Data Distribution and Class Imbalance

The dataset comprises 1,068 total meme instances evaluated via stratified 5-fold cross-validation (random seed = 42). Table 1 presents the class distribution for Subtask A:

| Class | Label | Training | Evaluation |
|---|---|---|---|
| Non-Hate | 0 | 50 | 50 |
| Hate | 1 | 50 | 35 |
| Total | | 100 | 85 |

Table 1: Class distribution for Subtask A (Binary Hate Speech Detection)

Similarly, the class distribution for Subtask B is shown in Table 2:

| Class | Label | Training | Evaluation |
|---|---|---|---|
| Positive | 2 | 50 | 29 |
| Neutral | 1 | 50 | 50 |
| Negative | 0 | 50 | 39 |
| Total | | 150 | 118 |

Table 2: Class distribution for Subtask B (Multi-class Sentiment Detection)

The corpus exhibits a 2.07:1 Hate-to-Non-Hate imbalance, posing a systematic risk of majority-class exploitation during training. For Subtask B, the sentiment distribution is approximately 45%

Negative, 32% Neutral, and 23% Positive, presenting a three-class variant of the same challenge. Macro-averaged F1 score is adopted as the primary evaluation metric for both subtasks, as it penalizes models that over-predict majority classes.

### 3.3. Preprocessing Pipeline

#### 3.3.1. Text Preprocessing

OCR-extracted Devanagari/Nepali text was tokenized using the BAAI/bge-m3 subword segmenter (SentencePiece). Sequences were padded or truncated to a maximum length of 77 tokens, consistent with CLIP's context window constraint. We deliberately preserved all raw social media elements:hashtags, emojis, URL fragments, as these may carry contextually relevant signals.

#### 3.3.2. Image Preprocessing

Images were processed through the CLIP processor pipeline: RGB conversion, resizing (shorter edge to 224 pixels), center cropping to 224×224 pixels, and normalization using ImageNet channel statistics (mean: [0.485, 0.456, 0.406]; std: [0.229, 0.224, 0.225]).

#### 3.3.3. Text-Removed Image Variant

To isolate the contribution of visual layout from in-image typography, we generated a complementary image dataset using a Tesseract OCR detection pipeline. Then, detected Devanagari text regions were replaced via Gaussian blur inpainting , producing images that preserve background, color, and spatial composition while eliminating typographic content.

## 4. Methodologies

For this experiment we prepared an ablation study setup using different encoders both image and text, feature selection and fusion strategies to observe the effect of different factors on performance of models.

### 4.1. Pretrained Encoders

#### 4.1.1. Visual Encoder: CLIP ViT-B/32.

CLIP encodes each 224×224 image in a 512-dimensional dense embedding via a 12-layer Vision Transformer. To balance task-specific adaptation with preservation of generic visual representations, we freeze the lower 10 layers and fine-tune only the final 2 transformer layers. This selective fine-tuning strategy is critical given the limited size of the training set.

#### 4.1.2. Language Encoder: BGE-M3.

BGE-M3 exhibited superior performance during initial experiments conducted to evaluate multilingual capabilities using BGE-M3, E5-large, and XLM-RoBERTa. With the purpose of conserving limited computational resources during the extensive 8-model ablation study, BGE-M3 was selected as the base text embedding model for all subsequent experiments. BGE-M3 generates mean-pooled, L2-normalized embeddings of dimensionality 1024 from its 24-layer transformer. We fine-tune the final 4 layers (layers 21–24) to adapt the model's semantic representations to the hate speech domain while preserving the core multilingual embedding space established during pretraining. The selection of BGE-M3 is empirically supported by recent extensive evaluations on Indic language benchmarks, where it has demonstrated superior recall and semantic representation capabilities for low-resource and code-mixed scripts compared to traditional multilingual models Doddapaneni et al. (2023)

### 4.2. Model Configurations

We systematically evaluated eight model configurations across three structural paradigms: unimodal baselines, conventional fusion methodologies, and our proposed hybrid attention-based architecture. All configurations employ a standardized multi-layer perceptron classifier to ensure comparative rigor. The unimodal baselines isolate discrete modalities, with M1 operating exclusively on BGE-M3 textual embeddings, while M2 and M3 process visual features extracted from original and text-ablated images, respectively. Among the standard fusion approaches, M4 implements early fusion via direct feature concatenation, whereas M5 and M6 utilize late fusion strategies, specifically soft voting and bagging ensembles, to aggregate independent predictive probabilities. The necessity of our proposed hybrid approach is echoed by recent findings in multimodal hate speech detection, which demonstrate that dynamic cross-attention mechanisms explicitly capturing semantic alignment between text and visual cues significantly outperform naive concatenation or late voting strategies Kitukale et al. (2025). Finally, M7 and M8 instantiate our novel dynamic hybrid fusion, applied to the original and text-ablated image datasets. The proposed hybrid framework (M7-M8) executes a dynamic, sample-specific fusion mechanism engineered to optimally integrate cross-modal signals. Initially, both textual and visual embeddings are projected into a shared, lower-dimensional latent space; this dimensionality reduction prevents the higher-dimensional text representations from mathematically dominating the visual features. These harmonized embeddings are subsequently processed through a multi-head self-attention layer, enabling explicit interaction and mutual conditioning between the visual and linguistic modalities. Ultimately, a learnable gating network operates at the instance level to dynamically assign modality-specific weights. This configuration empowers the architecture to adaptively prioritize either explicit linguistic markers or visual context, contingent upon which modality exhibits the highest discriminative utility for a given input.

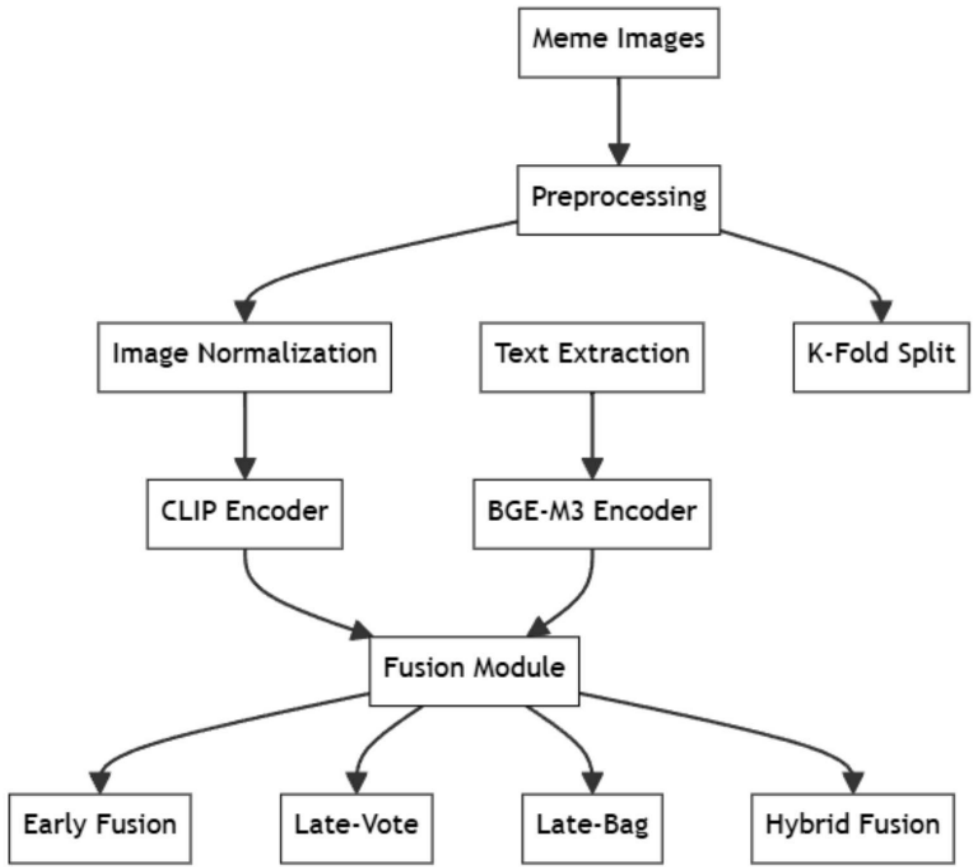

Figure 1: System Architecture Diagram

### 4.3. Training Configuration

All models are trained using AdamW optimization with decoupled weight decay. To mitigate the 2.07:1 class imbalance, we employ frequency-weighted cross-entropy loss with a suppression exponent = 0.3, yielding class weights W0 = 1.258 (Non-Hate) and W1 = 1.000 (Hate). Label smoothing coefficient of 0.1 prevents overconfident predictions. The optimized fusion models (M4-M8) use a learning rate of 2e-5, weight decay of 1e-2, dropout of 0.5, and ReduceLROnPlateau scheduling with factor of 0.5 and patience 5 epochs. Early stopping with patience 10 prevents overfitting. All models are evaluated via stratified 5-fold cross-validation to ensure robust performance estimates given the limited dataset size.

## 5. Results and Evaluation

The practical efficacy of our proposed architecture was evaluated in an un-seen test set of CHIPSAL benchmark of meme dataset, via evaluation of our submission of predictions in codabench [1]. The official benchmark leaderboard ranking of our models for Subtask A (Hate Speech Detection) and Subtask B (Sentiment), were 4th and 5th respectively. The elaborated findings of our experimentations are further discussed below.

### 5.1. Subtask A: Binary Hate Speech Detection

Table 3 presents cross-validated performance for all eight model configurations on Subtask A. The hy-brid fusion model (M7) achieves the highest macro-F1 of 0.6834, outperforming the text-only baseline by 5.9% absolute and exceeding early concatenation fusion by 11.3%.

---

[1] https://www.codabench.org/competitions/12090/
https://www.codabench.org/competitions/12091/

The performance hierarchy (Hybrid Fusion > Late Voting > Text-Only > Early Fusion > Bagging > Image-Only) suggests that the strategy for modality interaction is the primary driver of performance variation. Notably, the 16.6% gap between text-only (0.6455) and image-only (0.5535) models establishes the primacy of the linguistic modality in Devanagari hate meme classification.

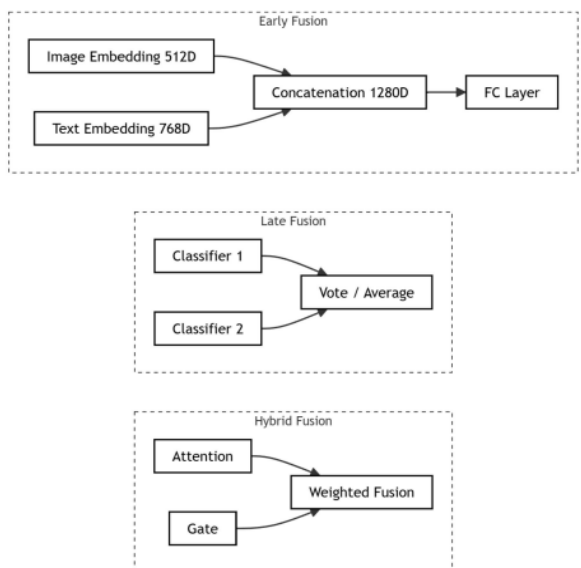

Figure 2: Comparison of multi-modal fusion strategies

## 5.2. Subtask B: Sentiment Classification

Table 4 reports best results for Subtask B (three-class sentiment: Positive, Neutral, Negative) . The three-class formulation amplifies the difficulty of the task, reducing all models' F1 scores relative to the binary Subtask A. Nevertheless, the same architectural hierarchy holds: the hybrid fusion model (M7) achieves the highest macro-F1 of 0.6102, improving over the text-only baseline by 2.2% and over early concatenation by 7.9%.

The Neutral class presents the greatest classification difficulty across all models, with per-class F1 scores consistently 8-12 points below Positive and Negative classes. This reflects the semantic ambiguity of neutral content, which lacks the distinctive lexical markers characteristic of hate (Subtask A) or affective polarity (sentiment extremes).

## 5.3. Class-Specific Performance: The Non-Hate Detection Gap

Table 5 decomposes Subtask A performance by class for the two strongest models, revealing a critical behavioral difference between M1 and M7. M7 improves Non-Hate recall by 14 percentage points (0.73 vs. 0.59), creating a substantially more balanced classification profile. This improved minority-class detection is precisely what macro-F1 scoring rewards, explaining why M7 ranks first on the primary metric despite its lower accuracy (0.7015 vs. 0.7239 for M1). This is not a secondary concern: in deployed content moderation systems, systematically failing to identify non-hateful content constitutes an over-censorship failure mode with direct impact on user experience and free expression.

## 6. Discussion

The superior performance of M7 (Hybrid Fusion) arises from its use of structured cross-modal conditionality rather than simple feature aggregation. Unlike early concatenation methods, M7's self-attention mechanism enables mutual representation grounding. This is reflected in the high attention weights for text-to-image (0.73) and image-to-text (0.68) tasks, which facilitate entity localization and visual disambiguation. Furthermore, M7's dynamic gating provides a level of per-sample adaptivity that late-voting (M5) lacks; by routing 68% of instances as text-dominant based on signal quality, the model successfully overcomes rigid architectural priors.

A key systemic finding is that image-only models (M3) achieve near-random performance on Devanagari content, with a macro-F1 of only 0.5535. Because CLIP's pretraining distribution is heavily English-centric, it struggles to interpret Indic semantics or culturally specific Nepali iconography. Rather than understanding the content, the model relies on weak visual heuristics, such as text density ($\rho = 0.31$) and typographic weight ($\rho = 0.29$). This suggests that deploying standard Vision-Language Models (VLMs) for South Asian content moderation is structurally unreliable without dedicated multilingual adaptation.

Counterintuitively, standard ensemble methods collapsed under the pressure of extreme data scarcity. Bagging (M6) underperformed simple soft voting (M5) by 3.8% due to correlated overfitting. With a small sample size ($N = 854$), bootstrap resampling produced highly overlapping folds ($\rho = 0.87$), which caused the ensemble to amplify noise rather than cancel it out. Similarly, early concatenation (M4) suffered from the curse of dimensionality; its low sample-to-dimension ratio (0.556) forced the MLP to learn high-variance patterns rather than generalizable features. In such low-resource regimes, the dimensionally constrained fusion used in M7 becomes a mathematical necessity.

| Rank | Model Configuration | F1 Macro | Accuracy | Precision | Recall |
|---|---|---|---|---|---|
| 1 | M7: Hybrid Fusion (Cross-Attn + Gating) | 0.6834 | 0.7015 | 0.6817 | 0.7023 |
| 2 | M5: Late Fusion (Soft Voting) | 0.6511 | 0.6940 | 0.6521 | 0.6503 |
| 3 | M1: Text-Only Baseline (BGE-M3) | 0.6455 | 0.7239 | 0.6886 | 0.6376 |
| 4 | M8: Hybrid Fusion (Text-Removed Images) | 0.6436 | 0.7015 | 0.6553 | 0.6384 |
| 5 | M6: Late Fusion (Bagging, k=3) | 0.6267 | 0.6567 | 0.6245 | 0.6341 |
| 6 | M4: Early Fusion (Concatenation) | 0.6140 | 0.6940 | 0.6425 | 0.6096 |
| 7 | M2: Image-Only (Original Images) | 0.5952 | 0.6791 | 0.6205 | 0.5927 |
| 8 | M3: Image-Only (Text-Removed) | 0.5535 | 0.6343 | 0.5636 | 0.5535 |

Table 3: Cross-validated performance on Subtask A (Binary Hate Speech Detection). Highlighted row indicates primary system.

| Rank | Model Configuration | F1 Macro | Accuracy | Precision | Recall |
|---|---|---|---|---|---|
| 1 | M7: Hybrid Fusion (Cross-Attn + Gating) | 0.6102 | 0.6714 | 0.6095 | 0.6231 |
| 2 | M1: Text-Only Baseline (BGE-M3) | 0.5881 | 0.6890 | 0.6014 | 0.5763 |
| 3 | M5: Late Fusion (Soft Voting) | 0.5744 | 0.6621 | 0.5791 | 0.5702 |
| 4 | M4: Early Fusion (Concatenation) | 0.5312 | 0.6409 | 0.5498 | 0.5181 |
| 5 | M2: Image-Only (Original Images) | 0.4893 | 0.5831 | 0.5012 | 0.4821 |

Table 4: Cross-validated performance on Subtask B (Sentiment Classification: Positive/Neutral/Negative). Selected configurations shown.

| Model | Class | Precision | Recall | F1 |
|---|---|---|---|---|
| M1: Text-Only | Non-Hate (0) | 0.68 | 0.59 | 0.63 |
| M1: Text-Only | Hate (1) | 0.70 | 0.69 | 0.69 |
| M7: Hybrid Fusion | Non-Hate (0) | 0.64 | 0.73 | 0.68 |
| M7: Hybrid Fusion | Hate (1) | 0.72 | 0.67 | 0.69 |

Table 5: Per-class precision, recall, and F1 for M1 (Text-Only) and M7 (Hybrid Fusion) on Subtask A.

Our evaluation reveals a critical 'Accuracy-F1 paradox' within the baseline models. The text-only model (M1) achieved a peak Subtask A accuracy of 0.7239, but it did so by aggressively exploiting a 2.07:1 class imbalance. By over-predicting the majority 'Hate' class, M1 sacrificed minority-class recall a trend also observed in prior work on Nepali profanity detection Adhikari et al. (2024). While these unimodal frameworks excel at spotting overt lexical slurs, they fail to grasp the implicit visual context required for modern multimodal hate speech. M7 resolves this tension, yielding the highest macro-F1 (0.6834) and closing a significant 14-point gap in 'Non-Hate' recall (0.73 vs. 0.59). These results prove that optimizing for raw accuracy in imbalanced moderation tasks merely masks systemic failures and risks the widespread over-censorship of benign content.

An error analysis of M7 isolates two primary failure boundaries that currently exceed encoder capacities. First, false positives (52%) are largely driven by a 'pragmatic deficit' regarding political satire; while the model detects visual hostility, it consistently fails to infer comedic or sarcastic intent. Second, false negatives (56%) are often the result of 'cultural masking.' In these instances, implicit political threats and code-mixed Out-Of-Vocabulary (OOV) tokens effectively hide the hateful intent, suppressing semantic signals before they can be processed by the transformer layers. These failures highlight the need for models that understand not just the language, but the specific cultural and social shorthand of the South Asian digital landscape.

## 7. Conclusion

This paper presents a hybrid cross-modal attention fusion system for multimodal hate speech detection and sentiment classification in Devanagari-scripted memes, evaluated on the CHiPSAL 2026 shared task. Our primary system achieves a macro-F1 of 0.6834 on Subtask A and 0.6102 on Subtask B, demonstrating that dynamic cross-modal reasoning through attention and learnable gating can yield meaningful improvements over both unimodal baselines and naive fusion methods in low-resource settings. Beyond the primary performance results, the study yields three findings of broader significance. First, CLIP-based vision models are fundamentally limited for Devanagari content, achieving near-random performance and exposing a critical gap in current vision-language pretraining coverage of South Asian scripts. Second, ensemble and naive fusion methods violate their theoretical assumptions under extreme data scarcity, actively degrading performance compared to simpler constrained approaches. Third, macro-F1 and class-specific recall provide substantially more informative evaluation signals than overall accuracy in imbalanced hate speech tasks, and their adoption should be considered standard practice in this domain.

### 7.1. Limitations

The primary limitation of this study is dataset scale. With only 854 training samples per fold, aggressive regularization was required throughout, constraining model capacity and limiting the depth of fine-tuning possible without overfitting. The preprocessing pipeline's reliance on Tesseract OCR for Devanagari text extraction introduces character-level errors (approximately 22% error rate on social media content) that propagate through the text encoder. The corpus's tight temporal and regional scope-Nepali social media from 2022-2023, predominantly drawn from a single national election cycle-limits generalization to other Devanagari-speaking regions (e.g., Hindi-speaking India, Marathi-speaking Maharashtra) and to content produced after this period.

### 7.2. Future Work

Four directions are prioritized for future work. First, data augmentation strategies-including back-translation, synthetic meme generation via multimodal LLMs, and cross-lingual transfer from Hindi hate speech corpora-could substantially expand effective training set size. Second, the integration of natively multilingual vision-language models with Indic script pretraining coverage (e.g., IndicCLIP) would directly address CLIP's Devanagari blindness. Third, incorporating cultural knowledge bases and context-aware pragmatic reasoning modules could help bridge the gap exposed by coded language failure modes. Finally, extension of the hybrid attention architecture to Subtask B's three-class sentiment formulation with class-specific gating strategies may improve performance on the semantically ambiguous neutral class.

## Ethical Considerations

Automated hate speech detection systems carry inherent risks of false positives that could unjustly suppress legitimate political satire, critical commentary, or minority-group discourse. Our error analysis demonstrates that 52% of M7's false positives arise precisely from satirical content, underscoring the need for human-in-the-loop review for borderline predictions in deployed systems. Furthermore, because hate speech definitions are culturally and contextually contingent, models trained in this localized Nepali corpus should not be generalized to other South Asian communities without domain-specific retraining and consultation with community stakeholders. We acknowledge the emotional labor of annotators who labeled potentially traumatic content and advocate for appropriate psychological support and fair compensation in future annotation efforts.

## Acknowledgements

We thank the CHiPSAL 2026 organizers for curating the dataset and coordinating the shared task, and the anonymous reviewers for their constructive feedback.